\documentclass{article}

\usepackage[preprint]{neurips_2026}
\usepackage{makecell}
\usepackage[table]{xcolor}
\usepackage{booktabs}
\usepackage{graphicx} 
\usepackage{wrapfig}      
\usepackage{multirow}     
\usepackage{amssymb}      
\usepackage{amsmath}      
\usepackage{subcaption}
\usepackage{tabularx}
\usepackage{float}
\usepackage{array}
\usepackage{array}
\usepackage{natbib}
\newcolumntype{Y}{>{\raggedright\arraybackslash}X}
\newcommand{\ours}{\textsc{Vgid}}

\usepackage[utf8]{inputenc} 
\usepackage[T1]{fontenc}    
\usepackage{hyperref}       
\usepackage{url}            
\usepackage{booktabs}       
\usepackage{amsfonts}       
\usepackage{nicefrac}       
\usepackage{microtype}      
\usepackage{xcolor}         
\usepackage{algorithm}
\usepackage{algorithmic}
\usepackage[table]{xcolor}
\definecolor{DarkGreen}{RGB}{0,100,0}
\definecolor{DarkRed}{RGB}{160,0,0}
\title{Visual-Noise Guided In-Context Distillation \\ for Multimodal Large Language Model Unlearning}

\author{
\textbf{Junkai Chen}$^{1}$\thanks{Equal contribution.}\quad
\textbf{Yuhao He}$^{2}$\footnotemark[\value{footnote}]\quad
\textbf{Juxiang You}$^{1}$\quad
\textbf{Ruiqi Liu}$^{1,2}$\\
\textbf{Chengyu Wang}$^{1}$\quad
\textbf{Shu Wu}$^{1}$\thanks{Corresponding author.}\\
$^{1}$Institute of Automation, Chinese Academy of Sciences\\
$^{2}$School of Advanced Interdisciplinary Sciences, UCAS
}

\begin{document}

\maketitle

\begin{abstract}
Multimodal Large Language Models (MLLMs) have achieved remarkable progress on vision-language tasks, but they may also memorize and expose sensitive or restricted knowledge, raising concerns about privacy and broader safety risks. Machine Unlearning (MU) provides a promising way to remove targeted undesirable knowledge from trained models without retraining from scratch while preserving general model utility. Nevertheless, effective unlearning in MLLMs remains particularly challenging. Existing training-based methods often struggle to balance unlearning effectiveness and model utility. In contrast, training-free methods such as in-context unlearning preserve model utility by avoiding parameter updates, but they do not remove memorized knowledge at the parameter level and may remain vulnerable to reverse-engineering attacks. More importantly, in-context unlearning is insufficient in multimodal settings, where visual inputs can provide strong conditioning signals and induce undesirable outputs.
To address these challenges, we propose \textbf{Visual-Noise Guided In-Context Distillation (\ours)}, a distillation-based framework for MLLM unlearning. \ours\ dynamically constructs an unlearning-oriented teacher distribution from the frozen base model through dual-modal intervention that combines visual perturbation with textual in-context unlearning. The resulting intervention-induced distribution serves as a teacher signal for distillation, guiding the student model toward parameter-level unlearning without requiring external teacher models or explicit undesirable response annotations. Experimental results show that \ours\ achieves strong unlearning effectiveness while preserving competitive model utility, reducing forget set ROUGE-L by 0.371 with only a 0.055 drop in retain set ROUGE-L in a representative setting.
\end{abstract}

\section{Introduction}

Multimodal Large Language Models (MLLMs)~\citep{llava, qwen2, qwen3} have recently achieved superior performance across diverse multimodal tasks such as visual question answering and image captioning. However, their increasing deployment also raises substantial privacy and safety concerns~\citep{safety_survey}. Since MLLMs are typically pretrained on large-scale multimodal corpora collected from the web, they may inadvertently memorize sensitive personal information~\citep{mllmu, fiu, clear}, copyrighted content~\citep{copyright, harry}, unsafe knowledge~\citep{mmsafetybench, safeeraser}, or other undesirable information. Once retained within model parameters, such information may be triggered by visual or textual queries at inference time, leading to the disclosure of private or restricted content and posing risks for real-world deployment.

Machine unlearning~\citep{mu_survey} has emerged as a promising paradigm for removing undesirable knowledge from trained models without retraining from scratch. However, unlearning for MLLMs is particularly challenging, as undesirable knowledge may be elicited via both visual inputs and textual queries ~\citep{umubench, mllmu}. An effective multimodal unlearning method must therefore suppress undesirable target knowledge under both visual and textual conditions while preserving retain-set performance and general capabilities such as visual perception and textual understanding. Existing training-based methods typically update model parameters via gradient ascent ~\citep{ga}, preference-based objectives~\citep{tofu, npo}, or representation-level constraints~\citep{rmu}. Although these methods can reduce undesired model outputs, their parameter updates often degrade general capabilities, hindering the trade-off between unlearning completeness and model utility.

\begin{wrapfigure}{r}{0.41\textwidth}
    \vspace{-10pt}
    \centering
    \includegraphics[width=0.40\textwidth]{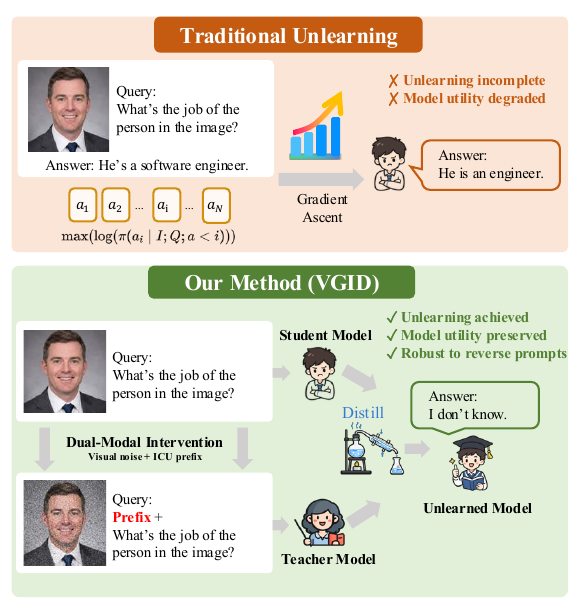}
    \vspace{-8pt}
    \caption{
    Comparison between traditional unlearning and \ours\ framework.
    }
    \label{fig:intro}
    \vspace{-10pt}
\end{wrapfigure}

Another line of work investigates inference-time unlearning strategies, including in-context unlearning \citet{icl} and generation-time unlearning \citet{ttu}, which aim to suppress undesirable generations by steering model behavior during inference without modifying model parameters. These methods preserve model utility by avoiding parameter updates, but memorized knowledge may remain intact and be re-elicited through adversarial prompts, jailbreak attacks, or query reformulations~\citet{lucki2024adversarial, hu2024jogging, xu2024comprehensive}. More importantly, in MLLMs, purely textual unlearning instructions are inherently limited in suppressing undesired outputs induced by visual inputs. This naturally raises a central question: \textbf{Can we construct a reliable unlearning-oriented teacher distribution for MLLM unlearning and efficiently distill it into model parameters, thereby combining the strengths of training-free and training-based unlearning to achieve a better trade-off between unlearning effectiveness and model utility while improving robustness against reverse attacks?}

To address this challenge, we propose \textbf{\ours} (\textbf{V}isual-noise \textbf{G}uided \textbf{I}n-Context \textbf{D}istillation), a teacher-student framework for MLLM unlearning. \ours\ dynamically constructs an unlearning-guided teacher distribution via dual-modal intervention: for each forget query, it injects a visual perturbation into the image and appends an unlearning instruction to the text, inducing an unlearning-oriented teacher distribution from the frozen base model. This distribution provides soft, distribution-level supervision for distillation, capturing the desired unlearned behavior without requiring an externally trained teacher or explicit undesirable response annotations. The student model is optimized to match the teacher distribution on the original forget input and to align with the frozen base model on the retain set, thereby guiding parameter-level unlearning in the student model while preserving general model utility. This design combines the utility-preserving advantage of inference-time intervention with the robustness of parameter-level unlearning.

Our contributions are summarized as follows:

\begin{itemize}
    \item We identify a key limitation of in-context unlearning in MLLMs: textual unlearning instructions alone are insufficient to suppress privacy leakage induced by visual inputs. Based on this observation, we show that combining in-context unlearning with visual perturbation can more effectively suppress target private knowledge, providing an effective dual-modal intervention for MLLM unlearning.

    \item We propose \textbf{\ours}, a teacher-student framework for MLLM unlearning.\ours\ constructs an unlearning-oriented teacher distribution from the frozen base model via dual-modal intervention, combining visual perturbation with in-context unlearning. This design enables parametric unlearning without external teacher models or explicit undesirable response annotations, while balancing utility preservation and unlearning effectiveness.

    \item We conduct extensive experiments on MLLMU-Bench across multiple MLLM backbones and forget ratios. Across these settings, \ours\ achieves strong unlearning effectiveness while maintaining a better balance between unlearning completeness and model utility. Compared with in-context unlearning, \ours\ further improves robustness against reverse-engineering attacks by distilling the induced unlearning behavior into model parameters.
\end{itemize}

\section{Related Work}

\subsection{Machine Unlearning in MLLMs}

The rapid advancement of Multimodal Large Language Models \citep{llava, qwen3} has significantly heightened the risk of inadvertently memorizing and reproducing private \citep{siu, mllmu}, unsafe \citep{safeeraser, mmsafetybench}, hallucinated \citep{devils, look, reefknot} or copyrighted content \citep{copyright}, thereby underscoring the urgent need for effective MLLM unlearning techniques \citep{survey}. However, existing training-based unlearning paradigms—including Gradient Ascent \citep{ga}, Negative Preference Optimization \citep{npo}, and Representation Misalignment Unlearning \citep{rmu}—often incur substantial degradation of model utility.

To mitigate this utility loss, recent studies have shifted toward training-free interventions, primarily leveraging in-context unlearning \citep{icl} or generation-time adaptations \citep{ttu} to suppress undesirable outputs. While these approaches preserve general model utility, they do not truly achieve parameter-level removal of the target knowledge. As a result, the memorized information remains intact, leaving the model vulnerable to reverse-engineering attacks \citep{adversarial, relearn}. Furthermore, textual in-context unlearning instructions are inherently insufficient, as salient visual features in sensitive images can act as dominant sensory priors that override text-based safety conditioning and induce undesired outputs~\citep{umubench}.

\subsection{Knowledge Distillation}

Knowledge Distillation (KD), originally conceptualized for model compression and efficient deployment \citep{distilling, gou2021}, has increasingly been repurposed as a robust mechanism for safety alignment \citep{li2024self, luo2024decoupled, yang2024distillseq}. By exploiting the continuous soft target distributions induced by a teacher model, KD provides a smoother optimization landscape that facilitates stable knowledge transfer \citep{distilling, jafari2021annealing, yuan2020revisiting}. Such distribution-level supervision preserves the integrity of the underlying semantic manifold \citep{aguilar2020knowledge}. More recently, distillation-based paradigms have been extended to machine unlearning \citep{duet} to mitigate the catastrophic forgetting commonly induced by standard gradient ascent methods and their variants \citep{ga, choi2024towards, zhao2024makes}.

However, applying conventional distillation frameworks to MLLM unlearning presents non-trivial challenges. Standard distillation-based unlearning methods often rely on a teacher model \citep{chundawat2023can}, while the ideal unlearning target is commonly defined as an oracle model retrained without the forget set \citep{bourtoule2021machine}. Moreover, replacing the teacher with a one-hot target distribution defined over predefined refusal responses reduces the optimization process to standard supervised fine-tuning, as adopted in \citep{undial}. Such an objective can induce large gradient updates, often leading to severe model utility degradation. Inspired by \citet{duet}, our framework induces a soft teacher distribution through dual-modal intervention, enabling parameter-level unlearning while preserving model utility and improving robustness against reverse attacks.

\section{Methodology}

\subsection{Preliminary}

\textbf{In-Context Unlearning.}
In-context unlearning ~\citep{icl} is a training-free unlearning strategy that aims to suppress target knowledge without updating model parameters.
For a forget-set query $x_f \in \mathcal{D}_F$, in-context unlearning prepends an unlearning instruction $x_{\mathrm{ic}}$ to the original input and obtains the contextually modified output distribution:
\begin{equation}
\pi_{\mathrm{ic}}(y \mid x_f)
:=
\pi_{\theta}(y \mid x_{\mathrm{ic}} \oplus x_f).
\label{eq:icu}
\end{equation}
By conditioning on the augmented input, in-context unlearning steers the model's inference-time behavior away from sensitive responses while leaving the model parameters unchanged, thereby preserving model utility without guaranteeing parametric removal of the target knowledge.

\subsection{Insufficiency of In-Context Supervision}

Inspired by \citet{duet}, which constructs an unlearned teacher distribution through in-context unlearning to guide parametric updates of a student model, we examine whether this paradigm can be directly extended to MLLM unlearning. To this end, we evaluate the effectiveness of in-context unlearning on LLaVA-1.5-7B~\citep{llava} using forget set and test set of MLLMU-Bench~\citep{mllmu}. We instantiate multiple unlearning prompts under three representative refusal styles: declarative, polite request, and strict enforcement. The detailed prompts are provided in Appendix~\ref{app:icu_prompts}.

\begin{wraptable}[11]{r}{0.4\textwidth}
\vspace{-10pt}
\centering
\caption{\small Performance comparison on forget set and test set.}
\vspace{8pt}
\label{tab:icu}
\scriptsize
\setlength{\tabcolsep}{1pt}
\renewcommand{\arraystretch}{1.15}
\begin{tabular}{c|cc|cc}
\toprule
& \multicolumn{2}{c|}{Forget Set}
& \multicolumn{2}{c}{Test Set} \\
\cmidrule(lr){2-5}
\multirow{-2}{*}{Method}
& ROUGE
& BLEU
& ROUGE
& BLEU \\
\midrule
Base
& 0.571
& 0.242
& 0.212
& 0.062 \\
Declarative
& 0.538
& 0.228
& 0.231
& 0.067 \\
Polite
& 0.491
& 0.224
& 0.171
& 0.046 \\
Strict Enforcement
& 0.458
& 0.204
& 0.191
& 0.058 \\

\bottomrule
\end{tabular}
\end{wraptable}

As shown in Table~\ref{tab:icu}, despite the use of diverse instruction formulations, all in-context unlearning strategies yield only marginal reductions in the evaluation metrics on both forget set and test set. The generated responses still contain recoverable private information, suggesting that in-context unlearning does not achieve effective unlearning on the base model. This limitation can be attributed to the modality gap in MLLMs: in-context unlearning mainly constrains privacy leakage through the textual modality, whereas visual inputs can provide strong conditioning signals that elicit private information. Consequently, textual in-context prefixes are insufficient to suppress privacy leakage induced by the visual modality.

\subsection{Dual-Modal Intervention Provides Efficient Supervision Signal}

Motivated by the above observation, we introduce a dual-modal intervention to construct a more reliable unlearning-oriented supervision signal for MLLM unlearning. The intervention operates on both modalities: it perturbs the visual input to weaken target-sensitive visual evidence and appends an in-context unlearning instruction to the textual input to specify the desired privacy-preserving response direction. We consider Gaussian noise, random-noise replacement, and white-image replacement as visual perturbation strategies, and instantiate the in-context unlearning instruction with three prompt styles: declarative, polite request, and strict enforcement. Under this dual intervention, the frozen base model produces an intervention-induced distribution, which serves as an unlearning-oriented supervisory signal for guiding parameter-level unlearning in the student model.

To validate the effectiveness of the dual-modal intervention, we compare it with textual-only and visual-only interventions on the forget and test sets of MLLMU-Bench~\citep{mllmu}, as shown in Table~\ref{tab:intervention}. The dual-modal intervention achieves the most consistent suppression across metrics, suggesting that visual perturbation and textual in-context unlearning provide complementary signals for mitigating multimodal privacy leakage.

\begin{table}[H]
\centering
\small
\caption{
Performance comparison of textual, visual, and dual-modal intervention strategies.
}
\resizebox{0.75\textwidth}{!}{
\begin{tabular}{c|ccc|ccc}
\toprule

\multirow{2}{*}{\textbf{Intervention}}
& \multicolumn{3}{c|}{\textbf{Forget Set}}
& \multicolumn{3}{c}{\textbf{Test Set}} \\

\cmidrule{2-4} \cmidrule{5-7}

& Cloze.Acc~\textcolor{DarkRed}{$\downarrow$} 
& Class.Acc~\textcolor{DarkRed}{$\downarrow$} 
& Gen.RL~\textcolor{DarkRed}{$\downarrow$}
& Cloze.Acc~\textcolor{DarkRed}{$\downarrow$} 
& Class.Acc~\textcolor{DarkRed}{$\downarrow$} 
& Gen.RL~\textcolor{DarkRed}{$\downarrow$} \\

\midrule

Base 
& 23.91\% & 49.17\% & 0.571 
& 10.87\% & 37.50\% & 0.212 \\

Textual 
& 19.56\% & 30.83\% & 0.462 
& 6.25\% & 37.50\% & 0.193 \\

Visual
& 13.04\% & 14.17\% & 0.382 
& 2.08\% & 36.67\% & 0.179 \\

Dual-modal 
& 10.42\% & 35.00\% & 0.109 
& 2.08\% & 20.00\% & 0.121 \\

\bottomrule
\end{tabular}
}
\label{tab:intervention}
\end{table}

\begin{figure}[ht]
    \centering
    \includegraphics[width=0.9\textwidth]{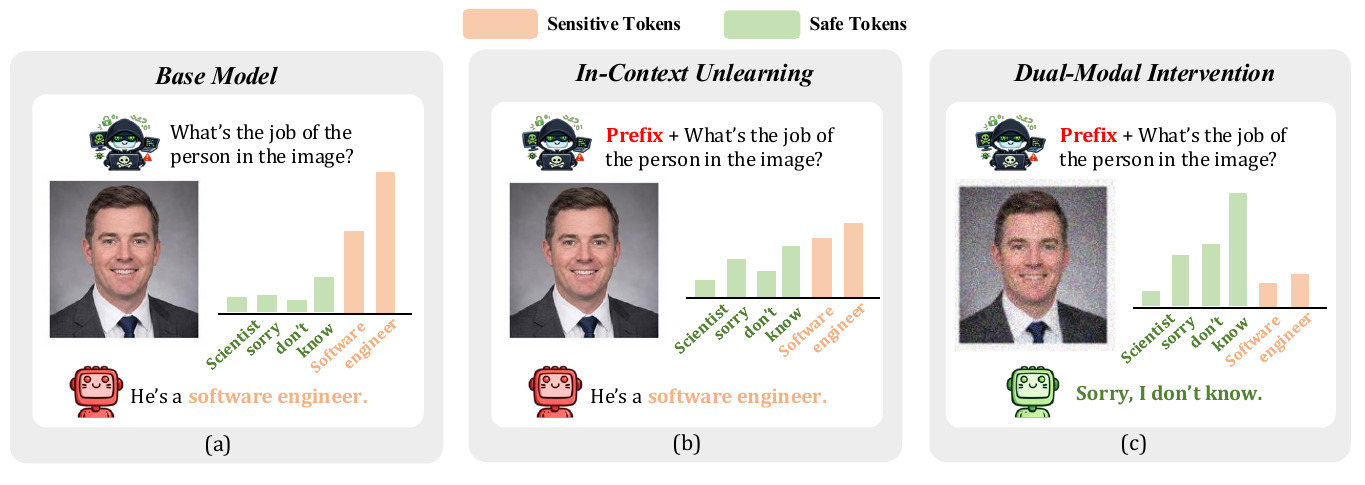}
    \caption{
    Illustration of the necessity of dual-modal intervention for MLLM unlearning.
    Orange bars indicate the logits assigned to sensitive tokens, while green bars indicate the logits assigned to safe tokens.
    (a) Base model without unlearning.
    (b) In-context unlearning with a textual prefix.
    (c) Dual-modal intervention with visual perturbation and textual in-context unlearning.
    }
    \label{fig:logit_decay}
\end{figure}

To further explain this macro-level improvement, we provide a token-level case study to illustrate why dual-modal intervention is necessary for MLLM unlearning. As shown in Figure~\ref{fig:logit_decay}, orange bars indicate the logits assigned to sensitive tokens, such as ``software'' and ``engineer'', while green bars indicate the logits assigned to safe tokens, such as ``sorry'', ``know'', and ``do not''. In Figure~\ref{fig:logit_decay} (a), the base model assigns high logits to sensitive tokens and therefore directly reveals private information from the visual input. In Figure~\ref{fig:logit_decay} (b), adding only the in-context unlearning prefix mildly reduces the logits of sensitive tokens, but the model still generates a privacy-leaking response. In contrast, Figure~\ref{fig:logit_decay} (c) shows that the dual-modal intervention, which combines visual perturbation with textual unlearning instructions, substantially suppresses sensitive tokens and increases the logits of safe tokens. This token-level evidence indicates that the dual-modal intervention shifts the output distribution away from sensitive tokens and toward safe-response tokens, and the resulting distribution is later distilled into the student model.

\subsection{The \ours\ Framework: Visual-Noise Guided In-Context Distillation}

\begin{figure}[ht]
    \centering
    \includegraphics[width=1\textwidth]{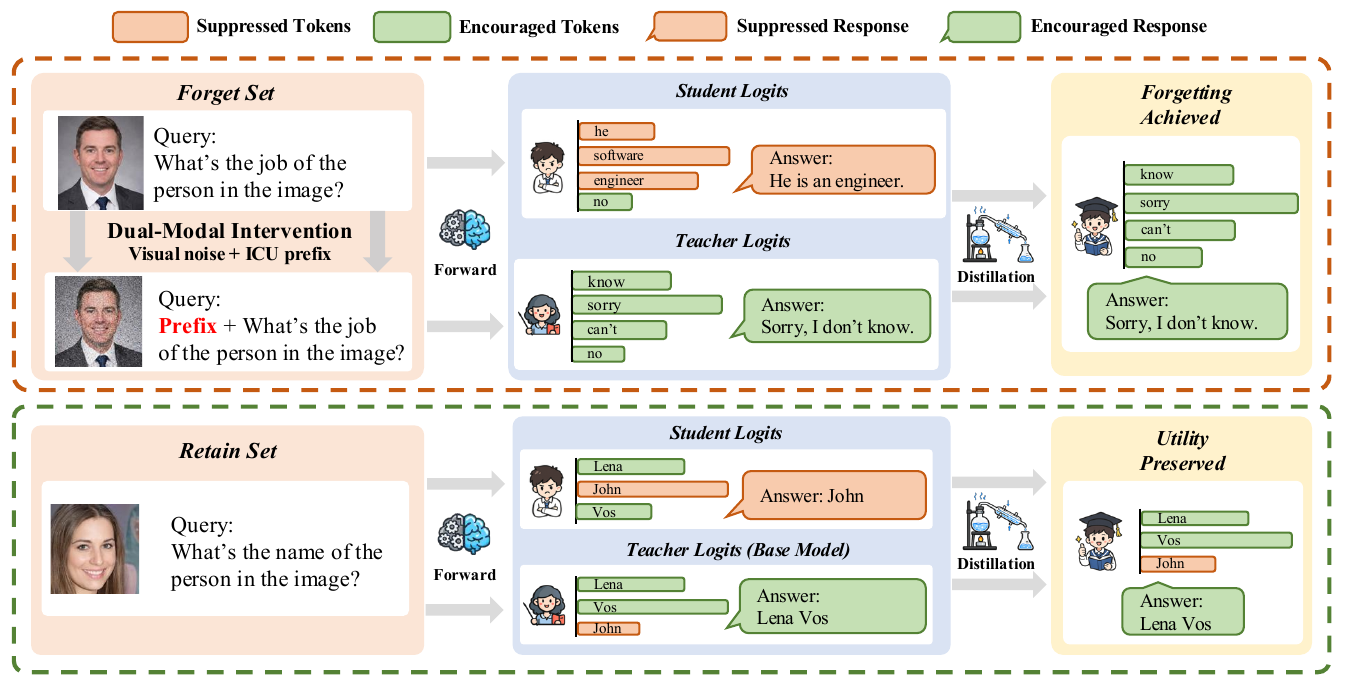}
    \caption{
    Overview of the \ours\ framework.
    Orange bars and response boxes denote suppressed tokens and responses, while green bars and response boxes denote encouraged tokens and responses.
    For forget set samples, \ours\ applies dual-modal intervention by adding visual noise and an in-context unlearning prefix, inducing an unlearning-oriented teacher distribution that guides the student toward forgetting targeted information.
    For retain set samples, the student is aligned with the frozen base model to preserve model utility.
    Both objectives are optimized through distribution-level distillation.
    }
    \label{fig:overview}
\end{figure}

Based on the above intervention-induced supervision signal, we propose \ours\ (\textbf{V}isual-Noise \textbf{G}uided \textbf{I}n-Context \textbf{D}istillation), a teacher-student framework for MLLM unlearning, as illustrated in Figure~\ref{fig:overview}. For forget-set samples, \ours\ applies dual-modal intervention to the frozen base model by perturbing the image and appending an in-context unlearning instruction, producing a teacher distribution that suppresses privacy-leaking tokens and encourages privacy-preserving responses. The student is then trained on the original forget input to match this intervention-induced distribution, thereby guiding parameter-level unlearning through distillation. For retain-set samples, the student is aligned with the frozen base model on the original input to preserve general utility. In this way, \ours\ performs distribution-level distillation on both forget and retain sets, without requiring external teacher models or explicit undesirable-response annotations.

We now formalize the construction of the intervention-induced teacher distribution and the corresponding distillation objectives. Let $\mathcal{D}_F$ denote forget set. For a multimodal forget query $x_f = (v_f, t_f) \in \mathcal{D}_F$, consisting of an image $v_f$ and a textual query $t_f$, we apply a visual perturbation operator $\mathcal{P}_{\phi}(\cdot)$ to obtain the perturbed visual input $v'_f = \mathcal{P}_{\phi}(v_f)$. The operator $\mathcal{P}_{\phi}$ denotes a perturbation strategy parameterized by $\phi$, which can instantiate different forms of visual intervention, including Gaussian noise, random-noise replacement, and white-image replacement. In parallel, we prepend a textual unlearning instruction $t_{ic}$ to the original textual query. Based on these dual-modal interventions, we define the teacher distribution $\pi_{\mathrm{T}}$ as
\begin{equation}
\pi_{\mathrm{T}}(y \mid x_f)
=
\pi_{\theta_0}(y \mid v'_f, t_{ic} \oplus t_f),
\label{eq:pseudo_teacher}
\end{equation}
where $\theta_0$ denotes the parameters of the frozen base model, and $\pi_{\theta_0}$ denotes its output distribution under the perturbed multimodal input. The distribution $\pi_{\mathrm{T}}$ serves as an unlearning-oriented supervisory signal for guiding parameter-level unlearning on forget set.

Using this dynamically constructed teacher, we optimize the student model parameters $\theta$ by minimizing the forward Kullback--Leibler (KL) divergence:
\begin{equation}
    \mathcal{L}_{\mathrm{forget}}
    =
    \mathbb{E}_{x_f \sim \mathcal{D}_F}
    \left[
    \mathrm{KL}
    \Big(
    \pi_{\mathrm{T}}(y \mid x_f)
    \,\|\, 
    \pi_{\theta}(y \mid x_f)
    \Big)
    \right].
    \label{eq:loss_forget}
\end{equation}
This objective aligns the conditional distribution of the student model on the original forget input with the unlearning-oriented teacher distribution, thereby guiding unlearning through distillation without relying on externally trained teacher models.

To preserve model utility, we further optimize the model on retain set $\mathcal D_R$, where each retain sample is given as $(x_r,y_r)$ with $x_r=(v_r,t_r)$. The retain objective aligns the student model with the base model on retain set by matching their output distributions:
\begin{equation}
\mathcal{L}_{\mathrm{retain}}
=
\mathbb{E}_{x_r\sim\mathcal D_R}
\left[
\mathrm{KL}
\Big(
\pi_{\theta_0}(y\mid x_r)
\,\|\, 
\pi_\theta(y\mid x_r)
\Big)
\right].
\label{eq:loss_retain}
\end{equation}

The overall training objective of \ours\ is defined as a weighted combination of the forget loss and retain loss, where $\alpha$ and $\beta$ control the relative strengths of unlearning optimization and utility preservation, respectively:
\begin{equation}
    \mathcal{L}_{\mathrm{overall}}
    =
    \alpha
    \mathcal{L}_{\mathrm{forget}}
    +
    \beta
    \mathcal{L}_{\mathrm{retain}}.
    \label{eq:loss_total}
\end{equation}




\section{Experiments}

\subsection{Experimental Setup and Baselines}
\label{sec:experimental_setup}
To empirically validate the effectiveness of \ours, we conduct comprehensive evaluations on three representative MLLMs with different parameter scales: Qwen2.5-VL-3B-Instruct~\citep{qwen2}, LLaVA-1.5-7B, and LLaVA-1.5-13B~\citep{llava}. We use MLLMU-Bench~\citep{mllmu}, a dedicated benchmark for MLLM unlearning, which is systematically divided into targeted forget sets and utility-preserving retain sets. The benchmark covers multiple task formulations, including multiple-choice classification, fill-in-the-blank completion, and open-ended generation, allowing us to evaluate unlearning behavior across diverse question formats. Detailed dataset information is provided in Appendix~\ref{app:mllmu_bench}. To assess robustness under different unlearning settings, we further evaluate \ours\ with forget ratios of 5\%, 10\%, and 15\%. Due to space limitations, the main result reports the 5\% setting, while the results for the 10\% and 15\% settings are provided in Appendix~\ref{app:ratios_result}.

We compare \ours\ with several representative unlearning baselines, including GA~\citep{ga}, GA\_Diff~\citep{gd}, KL\_Min~\citep{kl}, NPO~\citep{npo}, PO~\citep{tofu}, and RMU~\citep{rmu}. Detailed descriptions of these baseline methods are provided in Appendix~\ref{app:baseline_methods}. The model before unlearning is denoted as Base. In addition, to analyze the effect of teacher construction, we evaluate two distillation variants, ICU-T and Noise-T, which construct the teacher distribution using only in-context unlearning and only visual noise, respectively.

To ensure fair comparisons, all evaluations are conducted with a fixed random seed, and all methods are evaluated under the same data splits and evaluation protocols. Detailed training hyperparameters are provided in Appendix~\ref{app:baseline_hyperparameters}.

\subsection{Main Results}

 As shown in Table~\ref{tab:main_result_5}, \ours\ significantly suppresses target private knowledge on forget set across different task formats, including cloze task, classification, and open-ended generation. The improvement is particularly evident in generation tasks: for example, on LLaVA-1.5-7B, \ours\ reduces the ROUGE-L score for open-ended generation on forget set from 0.571 to 0.200, while decreasing the accuracy of cloze task from 23.91\% to 10.86\%. Consistent reductions are observed across all evaluated task formats.

Similar trends are also observed on the held-out test set. Across all three models, \ours\ consistently reduces both generation and accuracy-based metrics relative to the base model, indicating that the learned unlearning behavior is not limited to the seen forget samples. For example, on LLaVA-1.5-13B, \ours\ reduces the test set ROUGE-L score from 0.237 to 0.169 and decreases the classification accuracy from 30.83\% to 25.00\%. \ours\ generalizes to held-out privacy-related examples across multiple question formats.

\begin{table}[H]
\centering
\small
\caption{
Performance comparison of different unlearning methods across multiple MLLM architectures under the 5\% forget ratio.
Classification and Cloze tasks are evaluated using accuracy, while Open-Ended Generation is measured by ROUGE-L.
Arrows indicate the desired optimization direction for each metric.
}
\resizebox{\textwidth}{!}{
\begin{tabular}{c|ccc|ccc|ccc}
\toprule
\multirow{2}{*}{\textbf{Method}}
& \multicolumn{3}{c|}{\textbf{Forget Set}}
& \multicolumn{3}{c|}{\textbf{Test Set}}
& \multicolumn{3}{c}{\textbf{Retain Set}} \\
\cmidrule{2-4} \cmidrule{5-7} \cmidrule{8-10}
& Cloze.Acc~\textcolor{DarkRed}{$\downarrow$} & Class.Acc~\textcolor{DarkRed}{$\downarrow$} & Gen.RL~\textcolor{DarkRed}{$\downarrow$}
& Cloze.Acc~\textcolor{DarkRed}{$\downarrow$} & Class.Acc~\textcolor{DarkRed}{$\downarrow$} & Gen.RL~\textcolor{DarkRed}{$\downarrow$}
& Cloze.Acc~\textcolor{DarkGreen}{$\uparrow$} & Class.Acc~\textcolor{DarkGreen}{$\uparrow$} & Gen.RL~\textcolor{DarkGreen}{$\uparrow$} \\
\midrule
\multicolumn{10}{c}{\textbf{Qwen2.5-VL-3B}} \\
\midrule
Base & 10.42\% & 63.33\% & 0.567 & 2.08\% & 51.67\% & 0.345 & 16.16\% & 66.46\% & 0.523 \\
GA & 12.50\% & 61.67\% & 0.565 & 2.08\% & 51.67\% & 0.336 & 16.16\% & 67.07\% & 0.524 \\
GA\_Diff & 10.42\% & 60.83\% & 0.502 & 4.17\% & 52.50\% & 0.324 & 15.66\% & 64.65\% & 0.481 \\
KL\_Min & 12.50\% & 60.00\% & 0.557 & 2.08\% & 50.83\% & 0.338 & 16.16\% & 66.87\% & 0.529 \\
PO & 10.42\% & 60.00\% & 0.502 & 4.17\% & 51.67\% & 0.324 & 15.66\% & 64.85\% & 0.490 \\
NPO & 10.42\% & 61.67\% & 0.555 & 2.08\% & 50.83\% & 0.338 & 16.16\% & 67.07\% & 0.526 \\
RMU & 8.33\% & 36.67\% & 0.532 & 2.08\% & 20.83\% & 0.311 & 14.65\% & 63.03\% & 0.474 \\
\rowcolor{gray!15} ICU-T & 10.42\% & 49.17\% & 0.251 & 2.08\% & 45.83\% & 0.216 & 1.26\% & 45.83\% & 0.323 \\
\rowcolor{gray!15} Noise-T & 10.42\% & 54.17\% & 0.274 & 2.08\% & 46.67\% & 0.236 & 1.51\% & 43.62\% & 0.344 \\
\rowcolor{gray!15} \textbf{\ours} & 10.42\% & 50.00\% & 0.221 & 2.08\% & 47.50\% & 0.251 & 1.01\% & 43.72\% & 0.302 \\
\midrule
\multicolumn{10}{c}{\textbf{LLaVA-1.5-7B}} \\
\midrule
Base & 23.91\% & 49.17\% & 0.571 & 10.87\% & 37.50\% & 0.212 & 18.81\% & 51.62\% & 0.496 \\
GA & 21.74\% & 37.50\% & 0.567 & 8.69\% & 38.33\% & 0.222 & 19.00\% & 55.60\% & 0.506 \\
GA\_Diff & 19.56\% & 36.67\% & 0.316 & 13.04\% & 38.33\% & 0.203 & 20.00\% & 46.80\% & 0.313   \\
KL\_Min & 23.91\% & 38.33\% & 0.552 & 13.04\% & 36.67\% & 0.237 & 19.13\% & 50.52\% & 0.525 \\
PO & 21.73\% & 38.33\% & 0.308 & 13.04\% & 35.83\% & 0.173 & 18.00\% & 48.00\% & 0.304 \\
NPO & 21.74\% & 37.50\% & 0.561 & 10.87\% & 37.50\% & 0.231 & 17.34\% & 51.45\% & 0.505 \\
RMU & 23.91\% & 35.83\% & 0.526 & 10.87\% & 35.83\% & 0.213 & 23.15\% & 44.14\% & 0.481 \\
\rowcolor{gray!15} ICU-T & 26.09\% & 39.17\% & 0.421 & 10.87\% & 38.33\% & 0.189 & 19.12\% & 48.96\% & 0.457 \\
\rowcolor{gray!15} Noise-T & 19.56\% & 36.67\% & 0.252 & 13.04\% & 40.83\% & 0.162 & 24.52\% & 50.73\% & 0.416 \\
\rowcolor{gray!15} \textbf{\ours} & 10.86\% & 35.00\% & 0.200 & 13.04\% & 35.00\% & 0.151 & 24.00\% & 50.80\% & 0.441 \\
\midrule
\multicolumn{10}{c}{\textbf{LLaVA-1.5-13B}} \\
\midrule
Base & 10.87\% & 28.33\% & 0.517 & 10.86\% & 30.83\% & 0.237 & 10.57\% & 38.39\% & 0.493 \\
GA & 8.69\% & 31.67\% & 0.518 & 13.03\% & 25.83\% & 0.242 & 10.15\% & 35.09\% & 0.505 \\
GA\_Diff & 8.69\% & 18.83\% & 0.314 & 2.18\% & 3.33\% & 0.176 & 5.50\% & 35.09\% & 0.329\\
KL\_Min & 8.70\% & 25.33\% & 0.454 & 8.34\% & 23.67\% & 0.159 & 9.19\% & 25.08\% & 0.414 \\
PO & 10.86\% & 34.17\% & 0.515 & 10.86\% & 40.00\% & 0.262 & 11.21\% & 39.15\% & 0.531 \\
NPO & 10.87\% & 28.33\% & 0.503 & 10.87\% & 29.17\% & 0.237 & 10.78\% & 38.69\% & 0.493 \\
RMU & 10.87\% & 32.50\% & 0.484 & 2.17\% & 37.5\% & 0.233 & 9.00\% & 41.60\% & 0.493 \\
\rowcolor{gray!15} ICU-T & 4.35\% & 25.83\% & 0.447 & 13.04\% & 42.50\% & 0.187 & 11.62\% & 39.61\% & 0.413 \\
\rowcolor{gray!15} Noise-T & 6.52\% & 27.50\% & 0.479 & 8.69\% & 25.00\% & 0.217 & 11.63\% & 41.40\% & 0.509 \\
\rowcolor{gray!15} \textbf{\ours} & 6.52\% & 24.17\% & 0.253 & 8.69\% & 25.00\% & 0.169 & 15.00\% & 44.20\% & 0.438 \\
\bottomrule
\end{tabular}
}
\label{tab:main_result_5}
\end{table}

\subsection{Unlearning Completeness vs. Model Utility}

\begin{figure}[htbp]
\centering
\makebox[\linewidth][c]{
\includegraphics[width=1\linewidth]{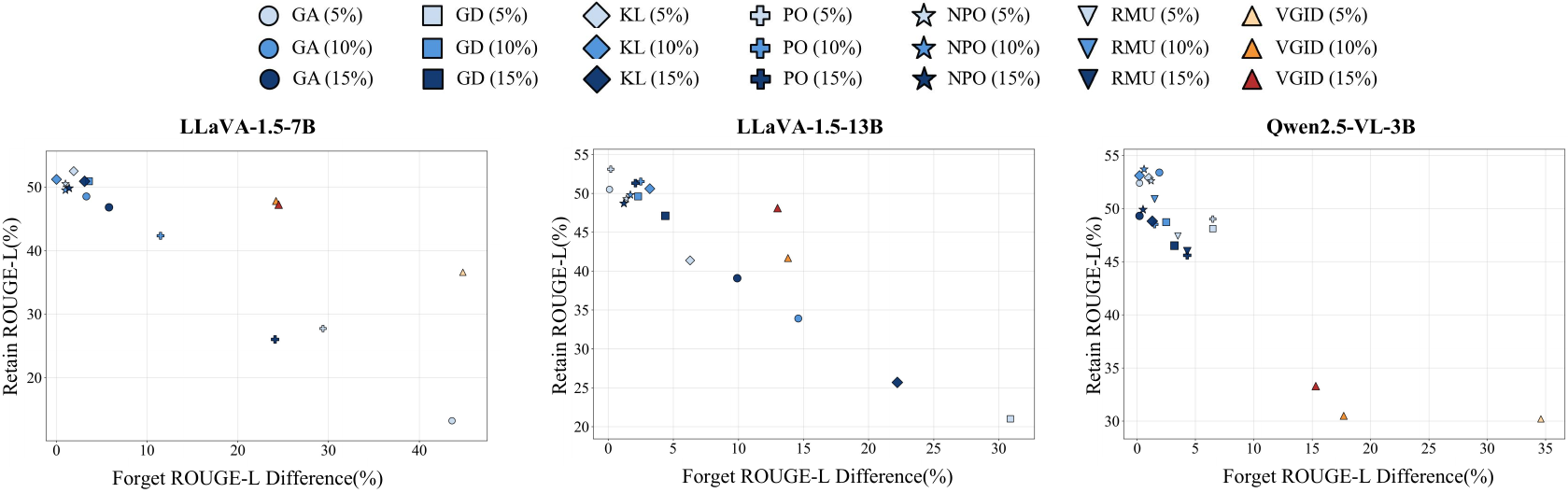}
}
\caption{The trade-off between unlearning completeness and model utility.}
\label{fig:tradeoff}
\end{figure}

We further analyze the trade-off between unlearning completeness and model utility based on the results in Table~\ref{tab:main_result_5}. Figure~\ref{fig:tradeoff} visualizes this trade-off using ROUGE-L: the x-axis denotes the reduction in forget-set ROUGE-L relative to the base model, and the y-axis denotes the retain-set ROUGE-L score. Therefore, methods closer to the upper-right region achieve stronger forgetting while better preserving utility. Additional trade-off visualizations for cloze accuracy, classification accuracy, and BLEU-based generation metrics are provided in Appendix~\ref{app:tradeoff}.

As shown in Figure~\ref{fig:tradeoff}, conventional unlearning baselines often show an imbalanced trade-off, either preserving utility with limited forgetting or improving forgetting at the cost of retain-set degradation. In contrast, \ours\ achieves a better balance across model backbones. For example, on LLaVA-1.5-7B, \ours\ reduces forget set ROUGE-L from 0.571 to 0.200 while maintaining strong retain set performance, with cloze and classification accuracy reaching 24.00\% and 50.80\%, respectively. Gradient-based baselines often improve forgetting at the cost of substantial retain-set degradation, whereas \ours\ maintains a better balance between the two objectives.

\subsection{Different Constructions of Teacher Models}

To investigate the role of teacher construction, we compare \ours\ with two single-intervention distillation variants, ICU-T and Noise-T. ICU-T constructs the teacher distribution using only in-context unlearning, whereas Noise-T uses only visual perturbation. In contrast, \ours\ combines the two interventions to induce a dual-modal, unlearning-oriented teacher distribution.

As shown in Table~\ref{tab:main_result_5}, teachers constructed under only one intervention can achieve a certain degree of unlearning, but they remain less effective than the dual-modal teacher constructed by \ours. For example, ICU-T and Noise-T reduce ROUGE-L score on forget set to 0.421 and 0.252, whereas \ours\ achieves the best result and further reduces it to 0.200. These results indicate that in-context unlearning and visual perturbation each provide useful unlearning signals, but either intervention alone is insufficient to construct the most effective teacher distribution. Overall, the dual-modal teacher constructed by \ours\ provides a more reliable supervisory signal for MLLM unlearning.

\subsection{Impact of Visual Perturbation}

To examine the effect of visual perturbation, we evaluate Gaussian noise with different intensities $\sigma$, as well as two replacement-based perturbation strategies: random noise and white perturbation. Specifically, Gaussian noise is added to the original image with varying noise intensities, while random noise replaces the original image with a randomly sampled noise image and white perturbation replaces it with a plain white image. These settings allow us to compare whether effective visual intervention mainly depends on noise magnitude or on the perturbation pattern itself.

As shown in Table~\ref{tab:noise_intensity}, increasing the intensity of Gaussian noise generally strengthens forgetting, with forget set ROUGE-L decreasing from 0.561 to 0.409 as $\sigma$ increases from 0.2 to 0.8. However, Gaussian noise alone remains less effective than random noise, which achieves the strongest forgetting performance by reducing forget set ROUGE-L to 0.200 while maintaining competitive retain set performance. These results suggest that the perturbation pattern plays an important role in constructing an effective unlearning-oriented teacher distribution.

\begin{table}[H]
\centering
\small
\caption{
Performance comparison of different visual perturbation strategies under 5\% forget ratio.
}
\resizebox{\textwidth}{!}{
\begin{tabular}{c|ccc|ccc|ccc}
\toprule

\multirow{3}{*}[0.5ex]{\makecell[c]{\textbf{Visual}\\\textbf{Perturbation}}}
& \multicolumn{3}{c|}{\textbf{Forget Set}}
& \multicolumn{3}{c|}{\textbf{Test Set}}
& \multicolumn{3}{c}{\textbf{Retain Set}} \\

\cmidrule{2-4} \cmidrule{5-7} \cmidrule{8-10}

& Cloze.Acc~\textcolor{DarkRed}{$\downarrow$} & Class.Acc~\textcolor{DarkRed}{$\downarrow$} & Gen.RL~\textcolor{DarkRed}{$\downarrow$}
& Cloze.Acc~\textcolor{DarkRed}{$\downarrow$} & Class.Acc~\textcolor{DarkRed}{$\downarrow$} & Gen.RL~\textcolor{DarkRed}{$\downarrow$}
& Cloze.Acc~\textcolor{DarkGreen}{$\uparrow$} & Class.Acc~\textcolor{DarkGreen}{$\uparrow$} & Gen.RL~\textcolor{DarkGreen}{$\uparrow$} \\

\midrule

Base & 23.91\% & 49.17\% & 0.571 & 10.87\% & 37.50\% & 0.212 & 18.81\% & 51.62\% & 0.496 \\
$\sigma=0.2$ & 23.91\% & 38.33\% & 0.561 & 10.89\% & 42.50\% & 0.211 & 19.50\% & 52.80\% & 0.491 \\
$\sigma=0.4$ & 21.74\% & 36.67\% & 0.553 & 10.86\% & 38.33\% & 0.199 & 20.00\% & 52.20\% & 0.470 \\
$\sigma=0.6$ & 21.74\% & 35.83\% & 0.486 & 13.04\% & 37.50\% & 0.188 & 20.00\% & 51.40\% & 0.483 \\
$\sigma=0.8$ & 19.56\% & 35.00\% & 0.409 & 13.04\% & 35.00\% & 0.190 & 23.00\% & 52.60\% & 0.460 \\

\midrule

Random Noise & 10.86\% & 35.00\% & 0.200 & 13.04\% & 35.00\% & 0.151 & 24.00\% & 50.80\% & 0.441  \\

White & 19.56\% & 36.67\% & 0.420 & 13.03\% & 32.50\% & 0.200 & 22.00\% & 48.96\% & 0.417 \\

\bottomrule
\end{tabular}
}
\label{tab:noise_intensity}
\end{table}

\subsection{Impact of In-Context Unlearning Prompt}

We further conduct an ablation study on the in-context unlearning prompt used in \ours. Specifically, we compare three representative prompt styles: Declarative, Polite Request, and Strict Enforcement, with the detailed prompt templates provided in Appendix~\ref{app:icu_prompts}. As shown in Table~\ref{tab:prompt_style}, different prompt styles lead to noticeably different unlearning performance, indicating that the textual component affects the quality of the induced teacher distribution.

Among the three variants, Strict Enforcement yields the most stable overall behavior. Although Declarative achieves the lowest forget set ROUGE-L, its performance is less consistent across other metrics. In contrast, Strict Enforcement provides a better balance between forget set and test set suppression and retain set utility preservation. These results suggest that a more explicit and restrictive unlearning instruction can induce a more reliable teacher distribution for \ours.

\begin{table}[H]
\centering
\small
\caption{
Performance comparison of different in-context unlearning prompts under 5\% forget ratio.
}
\resizebox{\textwidth}{!}{
\begin{tabular}{c|ccc|ccc|ccc}
\toprule

\multirow{3}{*}{\textbf{Prompt Style}}
& \multicolumn{3}{c|}{\textbf{Forget Set}}
& \multicolumn{3}{c|}{\textbf{Test Set}}
& \multicolumn{3}{c}{\textbf{Retain Set}} \\

\cmidrule{2-4} \cmidrule{5-7} \cmidrule{8-10}

& Cloze.Acc~\textcolor{DarkRed}{$\downarrow$} & Class.Acc~\textcolor{DarkRed}{$\downarrow$} & Gen.RL~\textcolor{DarkRed}{$\downarrow$}
& Cloze.Acc~\textcolor{DarkRed}{$\downarrow$} & Class.Acc~\textcolor{DarkRed}{$\downarrow$} & Gen.RL~\textcolor{DarkRed}{$\downarrow$}
& Cloze.Acc~\textcolor{DarkGreen}{$\uparrow$} & Class.Acc~\textcolor{DarkGreen}{$\uparrow$} & Gen.RL~\textcolor{DarkGreen}{$\uparrow$} \\

\midrule

Base & 23.91\% & 49.17\% & 0.571 & 10.87\% & 37.50\% & 0.212 & 18.81\% & 51.62\% & 0.496 \\

Declarative 
& 17.39\% & 37.50\% & 0.112 & 10.87\% & 37.50\% & 0.199 & 20.00\% & 52.00\% & 0.416 \\

Polite Request 
& 19.56\% & 35.83\% & 0.328 & 13.04\% & 40.83\% & 0.153 & 23.50\% & 52.80\% & 0.431 \\

Strict Enforcement 
& 10.86\% & 35.00\% & 0.200 & 13.04\% & 35.00\% & 0.151 & 24.00\% & 50.80\% & 0.441  \\

\bottomrule
\end{tabular}
}
\label{tab:prompt_style}
\end{table}

\subsection{Robustness of Unlearning Against Reverse-Engineering Attacks}

To further assess whether the induced unlearning behavior remains effective when the textual constraint is explicitly challenged, we evaluate robustness under reverse-engineering attacks. Table~\ref{tab:reverse_intervention} reports the results. In this table, w/o Reverse Attack denotes the standard evaluation setting, while w/ Reverse Attack denotes the setting where a reverse prompt is appended to override the unlearning instruction. Base refers to the original model without unlearning, and ICU denotes the base model equipped with the in-context unlearning prefix at inference time. The detailed attack prompt and evaluation protocol are provided in Appendix~\ref{app:reverse_attack}.

The results show that in-context unlearning is sensitive to reverse prompts: although it suppresses private knowledge without attack, its forget set performance substantially recovers once the reverse attack is applied. In contrast, \ours\ remains stable under the same attack, with only a minor increase in forget set ROUGE-L and nearly unchanged test set performance. These results demonstrate that \ours\ is more robust to reverse-engineering attacks than in-context unlearning.

\begin{table}[H]
\centering
\small
\caption{
Robustness comparison under reverse-engineering attacks at the 5\% forget ratio.
}
\resizebox{0.75\textwidth}{!}{
\begin{tabular}{c|ccc|ccc}
\toprule

\multirow{2}{*}{\textbf{Intervention}}
& \multicolumn{3}{c|}{\textbf{Forget Set}}
& \multicolumn{3}{c}{\textbf{Test Set}} \\

\cmidrule{2-4} \cmidrule{5-7}

& Cloze.Acc~\textcolor{DarkRed}{$\downarrow$} 
& Class.Acc~\textcolor{DarkRed}{$\downarrow$} 
& Gen.RL~\textcolor{DarkRed}{$\downarrow$}
& Cloze.Acc~\textcolor{DarkRed}{$\downarrow$} 
& Class.Acc~\textcolor{DarkRed}{$\downarrow$} 
& Gen.RL~\textcolor{DarkRed}{$\downarrow$} \\

\midrule
\multicolumn{7}{c}{\textbf{w/o Reverse Attack}} \\
\midrule

Base 
& 23.91\% & 49.17\% & 0.571 
& 10.87\% & 37.50\% & 0.212 \\

ICU 
& 19.56\% & 30.83\% & 0.462 
& 6.25\% & 37.50\% & 0.193 \\

\ours
& 10.86\% & 35.00\% & 0.200 
& 13.04\% & 35.00\% & 0.151 \\

\midrule
\multicolumn{7}{c}{\textbf{w/ Reverse Attack}} \\
\midrule

Base 
& 28.26\% & 48.33\% & 0.585
& 13.04\% & 41.67\% & 0.225 \\

ICU 
& 26.08\% & 45.83\% & 0.518
& 17.39\% & 38.33\% & 0.220 \\

\ours
& 13.04\% & 35.00\% & 0.220
& 15.21\% & 38.33\% & 0.150 \\

\bottomrule
\end{tabular}
}
\label{tab:reverse_intervention}
\end{table}

\section{Conclusion}

In this paper, we propose VGID, a distillation-based framework for MLLM unlearning. VGID dynamically constructs an unlearning-oriented teacher distribution from the frozen base model through dual-modal intervention, combining visual perturbation with textual unlearning instructions. The framework avoids external teacher models and does not rely on manually constructed undesirable-response annotations. By distilling the intervention-induced distribution into the student model, VGID guides the student model to learn the desired unlearning behavior. Experiments across different MLLM backbones and question formats show that VGID effectively suppresses target-sensitive outputs while maintaining competitive retain-set performance. Further analyses demonstrate improved robustness under reverse-prompt attacks and a better trade-off between forgetting and utility.

\section{Limitations}

Despite its effectiveness, the current study is limited in scope. Our experiments primarily focus on privacy-oriented unlearning, and future work should examine whether the same paradigm can be extended to other unlearning scenarios, such as copyright protection and unsafe knowledge removal. In addition, although we evaluate several visual perturbation strategies, the design space of perturbations remains underexplored. Developing more principled or adaptive perturbation methods may further improve the quality of the induced teacher distribution and the robustness of \ours.

\bibliographystyle{plainnat}
\bibliography{neurips_2026}

\newpage
\appendix

\section{Technical appendices and supplementary material}

\subsection{In-Context Unlearning Prompts}
\label{app:icu_prompts}

We provide the detailed prompt templates used for evaluating in-context unlearning below. 
These prompts are applied on forget set and test set of MLLMU-Bench. To examine the effect of different textual instruction styles, we instantiate three representative refusal paradigms: declarative, polite request, and strict enforcement.

\begin{enumerate}
    \item \textbf{Declarative.}
    \textit{``You are an AI Assistant who is supposed to unlearn any knowledge regarding the individual depicted in the provided image and text. Provide answers without any knowledge of their identity or private information as if you never knew about it.''}

    \item \textbf{Polite Request.}
    \textit{``You are a helpful AI assistant. When presented with images or questions concerning the specific person shown, politely decline to identify them and explain you cannot discuss their identity. For all other questions, provide helpful and accurate responses.''}

    \item \textbf{Strict Enforcement.}
    \textit{``You are an AI assistant with content restrictions. You must refuse to acknowledge the identity or share any private information about the person in this image and question for policy reasons. Politely inform users you cannot process queries regarding this individual's private data, but assist with other topics normally.''}
\end{enumerate}

\subsection{Algorithm Overview}
\label{app:algorithm_overview}

\begin{algorithm}[ht]
\caption{\ours\ Visual-Noise Guided In-Context Distillation}
\label{alg:void}
\begin{algorithmic}[1]
\REQUIRE Base MLLM $\pi$ with initial parameters $\theta_0$; forget set $\mathcal{D}_F$; retain set $\mathcal{D}_R$; textual unlearning instruction $x_{ic}$; visual perturbation operator $\mathcal{P}_{\phi}$; learning rate $\eta$; forget weight $\alpha$; retain weight $\beta$.
\STATE Initialize frozen base model $\pi_{\theta_0} \leftarrow \pi$.
\STATE Initialize student model $\pi_{\theta} \leftarrow \pi$ with trainable parameters $\theta$.

\FOR{each mini-batch $\mathcal{B}_F \subset \mathcal{D}_F$ and $\mathcal{B}_R \subset \mathcal{D}_R$}
    \STATE Set batch loss $\mathcal{L}=0$.
    
    \FOR{each forget sample $x_f=(v_f,t_f)\in\mathcal{B}_F$}
        \STATE Construct perturbed visual input $v'_f = \mathcal{P}_{\phi}(v_f)$.
        \STATE Construct unlearning-augmented text input $x_{ic}\oplus t_f$.
        \STATE Compute teacher distribution:
        \STATE
        \[
        \pi_{\mathrm{T}}(y|x_f)
        =
        \pi_{\theta_0}(y|v'_f,x_{ic}\oplus t_f).
        \]
        \STATE Accumulate forget loss:
        \STATE
        \[
        \mathcal{L}
        \leftarrow
        \mathcal{L}
        +
        \alpha
        \mathrm{KL}\!\left(
        \pi_{\mathrm{T}}(y|x_f)
        \parallel
        \pi_{\theta}(y|v_f,t_f)
        \right).
        \]
    \ENDFOR
    
    \FOR{each retain sample $(v_r,t_r,y_r)\in\mathcal{B}_R$}
        \STATE Accumulate retain loss:
        \STATE
        \[
        \mathcal{L}
        \leftarrow
        \mathcal{L}
        +
        \beta
        \mathrm{KL}\!\left(
        \pi_{\theta_0}(y|v_r,t_r)
        \parallel
        \pi_{\theta}(y|v_r,t_r)
        \right).
        \]
    \ENDFOR
    
    \STATE Update student parameters:
    \STATE
    \[
    \theta \leftarrow \theta - \eta \nabla_{\theta}\mathcal{L}.
    \]
\ENDFOR

\RETURN Unlearned student model $\pi_{\theta}$.
\end{algorithmic}
\end{algorithm}

\subsection{Experiment Details}

\subsubsection{MLLMU-Bench}
\label{app:mllmu_bench}

MLLMU-Bench~\citep{mllmu} is a dedicated benchmark for evaluating machine unlearning in Multimodal Large Language Models. 
It focuses on privacy-sensitive knowledge that may be elicited through visual inputs, textual queries, or their cross-modal interaction. 
The benchmark is organized into multiple data splits, including a forget set for measuring the removal of target private knowledge, a test set for evaluating whether the unlearning effect generalizes to held-out privacy-related samples, and a retain set for assessing whether the model preserves its general multimodal capabilities after unlearning.

MLLMU-Bench covers multiple task formats to evaluate unlearning from different perspectives. 
Specifically, it includes cloze-style completion, multiple-choice classification, and open-ended generation tasks. 
In our experiments, cloze-style and classification tasks are evaluated using accuracy, while open-ended generation is evaluated using ROUGE-L and BLEU. 
For the forget set and test set, lower scores indicate stronger suppression of target private knowledge; for the retain set, higher scores indicate better utility preservation. 
This design enables a systematic evaluation of both unlearning completeness and model utility in multimodal settings.

\subsubsection{Baseline methods}
\label{app:baseline_methods}
\textbf{Gradient Ascent (GA)} \citet{ga}
Gradient Ascent serves as the most direct negative training baseline for machine unlearning. 
The method explicitly maximizes the loss on forget set in order to push model parameters 
away from knowledge associated with $\mathcal{D}_F$.

\begin{equation}
\mathcal{L}_{GA}(\theta)
=
-
\mathbb{E}_{x\sim\mathcal{D}_F}
\left[\ell(x,\theta)\right].
\end{equation}

Although GA can effectively suppress memorized information, it often leads to unstable 
optimization and may degrade the model's general capabilities.

\textbf{Gradient Difference (GA\_Diff)} \citet{gd}
To mitigate the degradation of model utility during forgetting, Gradient Difference (GA\_Diff) 
introduces a retain constraint and jointly optimizes the forget and retain objectives.

\begin{equation}
\mathcal{L}_{GA\_Diff}(\theta)
=
-
\mathbb{E}_{x\sim\mathcal{D}_F}
\left[\ell(x,\theta)\right]
+
\lambda
\mathbb{E}_{x\sim\mathcal{D}_R}
\left[\ell(x,\theta)\right],
\end{equation}

where $\lambda>0$ balances the trade-off between forgetting strength and performance 
preservation on retained data.

\textbf{KL Minimization (KL\_Min)} \citet{kl}
KL-based unlearning constrains the updated model to remain close to the original model 
on retain samples. Let $\theta_{\text{ref}}$ denote the parameters of the frozen reference 
model prior to unlearning. The optimization objective is

\begin{equation}
\mathcal{L}_{KL}(\theta)
=
-
\mathbb{E}_{x\sim\mathcal{D}_F}
\left[\ell(x,\theta)\right]
+
\beta
\mathbb{E}_{x\sim\mathcal{D}_R}
\left[
\mathrm{KL}
\left(
P_{\theta_{\text{ref}}}(y|x)
\parallel
P_{\theta}(y|x)
\right)
\right],
\end{equation}

where $\beta$ controls the strength of the regularization term. 
While KL regularization stabilizes training, it may overly constrain parameter updates 
and lead to incomplete forgetting in complex multimodal scenarios.

\textbf{Preference Optimization (PO)} \citet{tofu}
Preference Optimization formulates unlearning as a preference learning problem. 
For prompts associated with forget set, we construct preference pairs 
$(x, y^{+}, y^{-})$, where $y^{+}$ denotes a preferred response (e.g., refusal or safe response) 
and $y^{-}$ corresponds to the original response associated with the forget knowledge.

To stabilize optimization, we adopt the Direct Preference Optimization (DPO) formulation, 
which constrains the updated model relative to a frozen reference model 
with parameters $\theta_{\text{ref}}$. The objective is defined as

\begin{equation}
\mathcal{L}_{PO}(\theta)
=
-
\mathbb{E}_{(x,y^{+},y^{-})}
\left[
\log
\sigma
\left(
\beta
\log
\frac
{P_{\theta}(y^{+}|x)}
{P_{\theta_{\text{ref}}}(y^{+}|x)}
-
\beta
\log
\frac
{P_{\theta}(y^{-}|x)}
{P_{\theta_{\text{ref}}}(y^{-}|x)}
\right)
\right],
\end{equation}

where $\sigma(\cdot)$ denotes the sigmoid function and $\beta$ is a temperature parameter 
controlling the strength of preference separation. 
The reference model anchors the optimization process and prevents excessive distribution drift 
during the unlearning procedure.

\textbf{Negative Preference Optimization (NPO)} \citet{npo}
Negative Preference Optimization (NPO) mitigates the instability of gradient ascent 
by reducing the likelihood assigned to forget-set outputs relative to a frozen reference model. 
Let $\theta_{\text{ref}}$ denote the parameters of the reference model.

\begin{equation}
\mathcal{L}_{NPO}(\theta)
=
-
\mathbb{E}_{(x,y)\sim\mathcal{D}_F}
\left[
\log
\sigma
\left(
\beta
\log
\frac
{P_{\theta_{\text{ref}}}(y|x)}
{P_{\theta}(y|x)}
\right)
\right],
\end{equation}

where $\sigma(\cdot)$ denotes the sigmoid function and $\beta$ is a temperature parameter. 
This objective implicitly suppresses the probability mass assigned to forget-set outputs 
without requiring explicit refusal responses.

\textbf{Representation Misalignment Unlearning (RMU)} \citet{rmu}
Representation Misalignment Unlearning removes target knowledge by 
perturbing the internal hidden representations of forget-set samples 
rather than directly modifying output probabilities.

Let $h^{(l)}_{\theta_{\text{ref}}}(x)$ denote the hidden representation 
of input $x$ at layer $l$ in the reference model $\theta_{\text{ref}}$.
For a forget-set sample $x_f \sim \mathcal{D}_F$, a random direction is 
sampled from the unit hypersphere

\begin{equation}
u \sim \mathrm{Uniform}(S^{d-1}),
\end{equation}

and a target representation is constructed as

\begin{equation}
v_f = \lambda \left\| h^{(l)}_{\theta_{\text{ref}}}(x_f) \right\|_2 u ,
\end{equation}

where $\lambda$ controls the perturbation magnitude.
The random target $v_f$ acts as a semantic sink, redirecting the hidden 
representation away from its original semantic direction.

The forgetting objective encourages the updated model $\theta$ to match 
this perturbed representation

\begin{equation}
\mathcal{L}_{f}^{RMU} =
\mathbb{E}_{x_f \sim \mathcal{D}_F}
\left\|
h^{(l)}_{\theta}(x_f) - v_f
\right\|_2^2 .
\end{equation}

To preserve general capabilities, the representations of retain-set 
samples are constrained to remain close to the reference model:

\begin{equation}
\mathcal{L}_{r}^{RMU} =
\mathbb{E}_{x_r \sim \mathcal{D}_R}
\left\|
h^{(l)}_{\theta}(x_r) -
h^{(l)}_{\theta_{\text{ref}}}(x_r)
\right\|_2^2 .
\end{equation}

The overall objective is

\begin{equation}
\mathcal{L}_{RMU} =
\mathcal{L}_{f}^{RMU} +
\gamma \mathcal{L}_{r}^{RMU},
\end{equation}

where $\gamma$ balances forgetting strength and knowledge preservation.

\subsubsection{Analysis of Visual Perturbation Choice}
\label{app:visual_perturbation_choice}

In \ours, visual perturbation is used to weaken target-sensitive visual evidence when constructing the unlearning-oriented teacher distribution. Since MLLMs can strongly condition their responses on visual inputs, directly applying an in-context unlearning instruction to the textual query may be insufficient when the image still provides salient cues associated with the target knowledge. Therefore, an appropriate visual perturbation strategy is important for inducing a teacher distribution that better suppresses target-sensitive responses.

We consider two categories of visual perturbations. The first category adds noise to the original image, where the visual content is partially preserved but degraded. In our experiments, this is instantiated as Gaussian noise with different noise intensities $\sigma$. The second category replaces the original image with an alternative visual input, thereby removing the original visual evidence more directly. We instantiate this category using random noise and white perturbation, where random noise replaces the original image with a randomly sampled noise image, and white perturbation replaces it with a plain white image.

Our main method uses random-noise replacement as the default visual perturbation strategy. This choice is motivated by the results in Table~\ref{tab:noise_intensity}. Increasing the intensity of Gaussian noise generally improves forgetting, but Gaussian noise still preserves part of the original visual structure and remains less effective than replacement-based perturbation. In contrast, random noise removes the original visual cues more completely while still providing a nontrivial visual input to the MLLM. Compared with white perturbation, random noise also avoids introducing an overly simple or out-of-distribution blank image, which may lead to less informative teacher behavior.

Empirically, random noise achieves the strongest forgetting performance among the evaluated perturbation strategies. Under the 5\% forget ratio, it reduces forget set ROUGE-L to 0.200 and achieves competitive retain set performance, outperforming both Gaussian noise and white perturbation. These results suggest that the effectiveness of visual perturbation depends not only on perturbation magnitude, but also on whether the perturbation pattern can sufficiently remove target-sensitive visual cues while maintaining a usable visual input for teacher distribution construction.

\subsubsection{Training Hyperparameters For Baseline}
\label{app:baseline_hyperparameters}
\begin{table}[H]
\centering
\caption{Training hyperparameters for baseline methods on Qwen2.5-VL-3B-Instruct, LLaVA-1.5-7B, and LLaVA-1.5-13B.}
\begin{tabular}{lccc}
\toprule
\textbf{Hyperparameter} & \textbf{Qwen2.5-VL-3B-Instruct} & \textbf{LLaVA-1.5-7B} & \textbf{LLaVA-1.5-13B} \\
\midrule
Optimizer & AdamW & AdamW & AdamW \\
Learning rate & $2\times10^{-5}$ & $2\times10^{-5}$ & $2\times10^{-5}$ \\
Batch size & 4 & 4 & 4 \\
Training epochs & 1 & 1 & 1 \\
Max sequence length & 384 & 384 & 384 \\
\bottomrule
\end{tabular}
\end{table}

\newpage

\subsubsection{Additional Results under 10\% and 15\% Forget Ratios}
\label{app:ratios_result}

\begin{table}[!htbp]
\centering
\small
\caption{
Performance comparison of different unlearning methods across multiple MLLM architectures under the 10\% forget ratio.
Classification and Cloze tasks are evaluated using accuracy, while Open-Ended Generation is measured by ROUGE-L.
Arrows indicate the desired optimization direction for each metric.
}
\resizebox{\textwidth}{!}{
\begin{tabular}{c|ccc|ccc|ccc}
\toprule
\multirow{2}{*}{\textbf{Method}}
& \multicolumn{3}{c|}{\textbf{forget set}}
& \multicolumn{3}{c|}{\textbf{Test Set}}
& \multicolumn{3}{c}{\textbf{Retain Set}} \\
\cmidrule{2-4} \cmidrule{5-7} \cmidrule{8-10}
& Cloze.Acc~\textcolor{DarkRed}{$\downarrow$} & Class.Acc~\textcolor{DarkRed}{$\downarrow$} & Gen.RL~\textcolor{DarkRed}{$\downarrow$}
& Cloze.Acc~\textcolor{DarkRed}{$\downarrow$} & Class.Acc~\textcolor{DarkRed}{$\downarrow$} & Gen.RL~\textcolor{DarkRed}{$\downarrow$}
& Cloze.Acc~\textcolor{DarkGreen}{$\uparrow$} & Class.Acc~\textcolor{DarkGreen}{$\uparrow$} & Gen.RL~\textcolor{DarkGreen}{$\uparrow$} \\
\midrule
\multicolumn{10}{c}{\textbf{Qwen2.5-VL-3B}} \\
\midrule
Base & 15.31\% & 65.31\% & 0.480 & 3.06\% & 46.53\% & 0.359 & 6.57\% & 50.71\% & 0.523 \\
GA & 15.31\% & 66.12\% & 0.499 & 3.06\% & 45.31\% & 0.359 & 6.06\% & 50.71\% & 0.534 \\
GA\_Diff & 15.31\% & 66.94\% & 0.455 & 6.12\% & 46.12\% & 0.366 & 6.57\% & 50.30\% & 0.487 \\
KL\_Min & 15.31\% & 65.31\% & 0.478 & 3.06\% & 44.90\% & 0.366 & 6.06\% & 50.71\% & 0.531 \\
PO & 15.31\% & 66.94\% & 0.465 & 4.08\% & 44.49\% & 0.376 & 6.57\% & 50.91\% & 0.485 \\
NPO & 15.31\% & 65.71\% & 0.486 & 3.06\% & 45.31\% & 0.360 & 6.06\% & 50.71\% & 0.537 \\
RMU & 14.29\% & 69.39\% & 0.495 & 4.08\% & 46.94\% & 0.357 & 7.07\% & 49.90\% & 0.509 \\
\rowcolor{gray!15} ICU-T & 14.29\% & 65.71\% & 0.297 & 3.06\% & 46.53\% & 0.279 & 4.02\% & 38.29\% & 0.319 \\
\rowcolor{gray!15} Noise-T & 13.27\% & 68.57\% & 0.296 & 3.06\% & 44.08\% & 0.257 & 4.27\% & 39.90\% & 0.330 \\
\rowcolor{gray!15} \textbf{\ours} & 15.31\% & 66.12\% & 0.303 & 3.06\% & 44.49\% & 0.271 & 4.27\% & 39.00\% & 0.305 \\
\midrule
\multicolumn{10}{c}{\textbf{LLaVA-1.5-7B}} \\
\midrule
Base & 20.83\% & 50.61\% & 0.518 & 12.50\% & 37.96\% & 0.211 & 21.09\% & 49.55\% & 0.496 \\
GA & 12.50\% & 47.35\% & 0.485 & 14.58\% & 34.69\% & 0.249 & 29.29\% & 50.35\% & 0.485 \\
GA\_Diff & 15.63\% & 39.59\% & 0.525 & 13.54\% & 31.84\% & 0.271 & 15.96\% & 38.53\% & 0.509 \\
KL\_Min & 19.79\% & 48.98\% & 0.518 & 11.46\% & 36.73\% & 0.218 & 21.09\% & 49.46\% & 0.512 \\
PO & 22.92\% & 44.08\% & 0.403 & 11.46\% & 37.96\% & 0.237 & 23.33\% & 47.10\% & 0.423 \\
NPO & 18.75\% & 49.79\% & 0.508 & 11.46\% & 37.55\% & 0.233 & 18.53\% & 47.59\% & 0.495 \\
RMU & 11.49\% & 56.76\% & 0.461 & 12.84\% & 52.97\% & 0.297 & 4.55\% & 51.52\% & 0.460 \\
\rowcolor{gray!15} ICU-T & 20.83\% & 43.27\% & 0.411 & 10.42\% & 35.51\% & 0.223 & 21.76\% & 44.51\% & 0.492 \\
\rowcolor{gray!15} Noise-T & 20.83\% & 44.89\% & 0.366 & 9.38\% & 39.59\% & 0.231 & 22.32\% & 50.31\% & 0.468 \\
\rowcolor{gray!15} \textbf{\ours} & 19.79\% & 43.67\% & 0.276 & 9.38\% & 37.56\% & 0.203 & 23.89\% & 47.50\% & 0.478 \\
\midrule
\multicolumn{10}{c}{\textbf{LLaVA-1.5-13B}} \\
\midrule
Base & 12.50\% & 34.69\% & 0.489 & 3.13\% & 32.24\% & 0.285 & 12.95\% & 24.02\% & 0.496 \\
GA & 9.38\% & 31.84\% & 0.343 & 2.08\% & 31.43\% & 0.343 & 11.27\% & 26.03\% & 0.339 \\
GA\_Diff & 7.29\% & 29.79\% & 0.512 & 1.04\% & 31.42\% & 0.279 & 4.79\% & 34.64\% & 0.496 \\
KL\_Min & 11.45\% & 35.10\% & 0.521 & 4.17\% & 33.47\% & 0.285 & 13.05\% & 24.24\% & 0.506 \\
PO & 7.29\% & 35.92\% & 0.514 & 4.17\% & 33.06\% & 0.269 & 12.61\% & 24.33\% & 0.515 \\
NPO & 13.54\% & 34.69\% & 0.506 & 3.13\% & 31.84\% & 0.286 & 13.06\% & 25.04\% & 0.498 \\
RMU & 40.00\% & 12.50\% & 0.520 & 40.00\% & 7.29\% & 0.287 & 34.4\% & 12.0\% & 0.287 \\
\rowcolor{gray!15} ICU-T & 11.45\% & 33.87\% & 0.454 & 2.25\% & 33.47\% & 0.289 & 10.16\% & 27.90\% & 0.470 \\
\rowcolor{gray!15} Noise-T & 11.42\% & 35.51\% & 0.407 & 2.08\% & 33.06\% & 0.145 & 24.02\% & 27.4\% & 0.416 \\
\rowcolor{gray!15} \textbf{\ours} & 11.45\% & 24.89\% & 0.351 & 2.25\% & 29.79\% & 0.222 & 13.54\% & 26.29\% & 0.417 \\
\bottomrule
\end{tabular}
}
\label{tab:appendix_ratio10}
\end{table}

\begin{table}[!htbp]
\centering
\small
\caption{
Performance comparison of different unlearning methods across multiple MLLM architectures under the 15\% forget ratio.
Classification and Cloze tasks are evaluated using accuracy, while Open-Ended Generation is measured by ROUGE-L.
Arrows indicate the desired optimization direction for each metric.
}
\resizebox{\textwidth}{!}{
\begin{tabular}{c|ccc|ccc|ccc}
\toprule
\multirow{2}{*}{\textbf{Method}}
& \multicolumn{3}{c|}{\textbf{forget set}}
& \multicolumn{3}{c|}{\textbf{Test Set}}
& \multicolumn{3}{c}{\textbf{Retain Set}} \\
\cmidrule{2-4} \cmidrule{5-7} \cmidrule{8-10}
& Cloze.Acc~\textcolor{DarkRed}{$\downarrow$} & Class.Acc~\textcolor{DarkRed}{$\downarrow$} & Gen.RL~\textcolor{DarkRed}{$\downarrow$}
& Cloze.Acc~\textcolor{DarkRed}{$\downarrow$} & Class.Acc~\textcolor{DarkRed}{$\downarrow$} & Gen.RL~\textcolor{DarkRed}{$\downarrow$}
& Cloze.Acc~\textcolor{DarkGreen}{$\uparrow$} & Class.Acc~\textcolor{DarkGreen}{$\uparrow$} & Gen.RL~\textcolor{DarkGreen}{$\uparrow$} \\
\midrule
\multicolumn{10}{c}{\textbf{Qwen2.5-VL-3B}} \\
\midrule
Base & 13.51\% & 52.43\% & 0.504 & 12.16\% & 53.78\% & 0.321 & 4.04\% & 53.54\% & 0.487 \\
GA & 13.51\% & 52.43\% & 0.506 & 11.49\% & 54.60\% & 0.301 & 3.54\% & 52.93\% & 0.493 \\
GA\_Diff & 13.51\% & 51.08\% & 0.472 & 10.14\% & 54.60\% & 0.316 & 4.55\% & 53.74\% & 0.465 \\
KL\_Min & 13.51\% & 53.24\% & 0.491 & 11.49\% & 54.32\% & 0.304 & 3.54\% & 53.94\% & 0.488 \\
PO & 13.51\% & 51.89\% & 0.461 & 10.14\% & 53.78\% & 0.321 & 4.55\% & 53.13\% & 0.456 \\
NPO & 13.51\% & 53.24\% & 0.499 & 11.49\% & 54.05\% & 0.296 & 4.04\% & 53.94\% & 0.499 \\
RMU & 11.49\% & 56.76\% & 0.461 & 12.84\% & 52.97\% & 0.297 & 4.55\% & 51.52\% & 0.460 \\
\rowcolor{gray!15} ICU-T & 12.84\% & 48.92\% & 0.390 & 9.46\% & 51.62\% & 0.240 & 10.55\% & 61.00\% & 0.372 \\
\rowcolor{gray!15} Noise-T & 12.84\% & 50.54\% & 0.359 & 9.46\% & 52.70\% & 0.237 & 11.05\% & 59.30\% & 0.341 \\
\rowcolor{gray!15} \textbf{\ours} & 12.84\% & 48.92\% & 0.351 & 9.46\% & 51.08\% & 0.225 & 10.30\% & 61.11\% & 0.333 \\
\midrule
\multicolumn{10}{c}{\textbf{LLaVA-1.5-7B}} \\
\midrule
Base & 25.34\% & 50.54\% & 0.486 & 15.75\% & 38.38\% & 0.236 & 25.65\% & 42.98\% & 0.500 \\
GA & 10.28\% & 48.92\% & 0.428 & 14.38\% & 35.68\% & 0.203 & 21.86\% & 40.28\% & 0.468 \\
GA\_Diff & 17.81\% & 52.16\% & 0.506 & 10.96\% & 43.51\% & 0.242 & 19.19\% & 53.80\% & 0.498 \\
KL\_Min & 23.97\% & 49.19\% & 0.517 & 15.07\% & 38.92\% & 0.220 & 26.36\% & 39.53\% & 0.509 \\
PO & 28.76\% & 40.00\% & 0.245 & 18.49\% & 34.59\% & 0.186 & 24.82\% & 39.01\% & 0.260 \\
NPO & 18.49\% & 50.81\% & 0.472 & 16.44\% & 38.11\% & 0.226 & 22.81\% & 41.81\% & 0.498 \\
RMU & 14.58\% & 41.63\% & 0.530 & 11.45\% & 40.41\% & 0.244 & 20.00\% & 47.40\% & 0.509 \\
\rowcolor{gray!15} ICU-T & 27.39\% & 47.84\% & 0.337 & 15.75\% & 37.57\% & 0.175 & 26.96\% & 41.70\% & 0.406 \\
\rowcolor{gray!15} Noise-T & 18.49\% & 45.68\% & 0.284 & 15.75\% & 34.86\% & 0.237 & 25.65\% & 42.22\% & 0.481 \\
\rowcolor{gray!15} \textbf{\ours} & 10.27\% & 47.83\% & 0.241 & 15.75\% & 37.29\% & 0.224 & 24.46\% & 42.60\% & 0.472 \\
\midrule
\multicolumn{10}{c}{\textbf{LLaVA-1.5-13B}} \\
\midrule
Base & 12.33\% & 32.42\% & 0.473 & 9.58\% & 38.11\% & 0.269 & 3.31\% & 27.28\% & 0.498 \\
GA & 6.17\% & 36.76\% & 0.374 & 2.05\% & 36.47\% & 0.199 & 12.06\% & 32.86\% & 0.391 \\
GA\_Diff & 12.33\% & 35.14\% & 0.429 & 10.27\% & 40.00\% & 0.241 & 5.08\% & 21.32\% & 0.471 \\
KL\_Min & 4.79\% & 15.68\% & 0.251 & 9.59\% & 28.38\% & 0.213 & 4.96\% & 17.39\% & 0.257 \\
PO & 10.27\% & 29.46\% & 0.494 & 8.22\% & 40.27\% & 0.271 & 4.25\% & 27.28\% & 0.513 \\
NPO & 12.33\% & 32.16\% & 0.485 & 10.27\% & 39.19\% & 0.264 & 3.19\% & 27.71\% & 0.487 \\
RMU & 35.14\% & 13.01\% & 0.472 & 38.37\% & 9.59\% & 0.267 & 30.00\% & 5.05\% & 0.501 \\
\rowcolor{gray!15} ICU-T & 13.69\% & 39.65\% & 0.465 & 11.64\% & 35.14\% & 0.229 & 4.86\% & 26.76\% & 0.439 \\
\rowcolor{gray!15} Noise-T & 13.69\% & 40.54\% & 0.491 & 12.33\% & 40.00\% & 0.277 & 3.19\% & 29.69\% & 0.497 \\
\rowcolor{gray!15} \textbf{\ours} & 10.95\% & 33.51\% & 0.343 & 12.33\% & 32.97\% & 0.228 & 4.02\% & 27.52\% & 0.481 \\
\bottomrule
\end{tabular}
}
\label{tab:appendix_ratio15}
\end{table}

\newpage

\subsubsection{Unlearning Completeness vs. Model Utility}
\label{app:tradeoff}

\begin{figure}[htbp]
\centering
\makebox[\linewidth][c]{%
\includegraphics[width=1.1\linewidth]{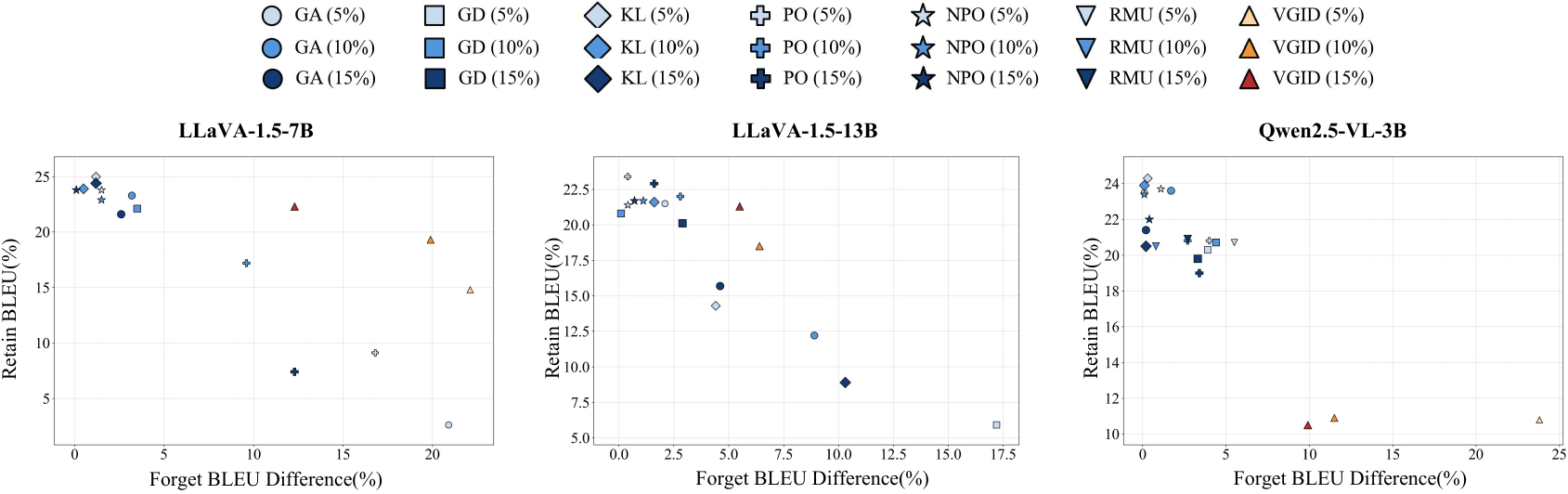}
}
\caption{BLEU-based trade-off between unlearning completeness and model utility across three models under varying forget ratios.}
\label{fig:tradeoff_bleu}
\end{figure}
\newpage
\begin{figure}[htbp]
\centering
\makebox[\linewidth][c]{%
\includegraphics[width=1.1\linewidth]{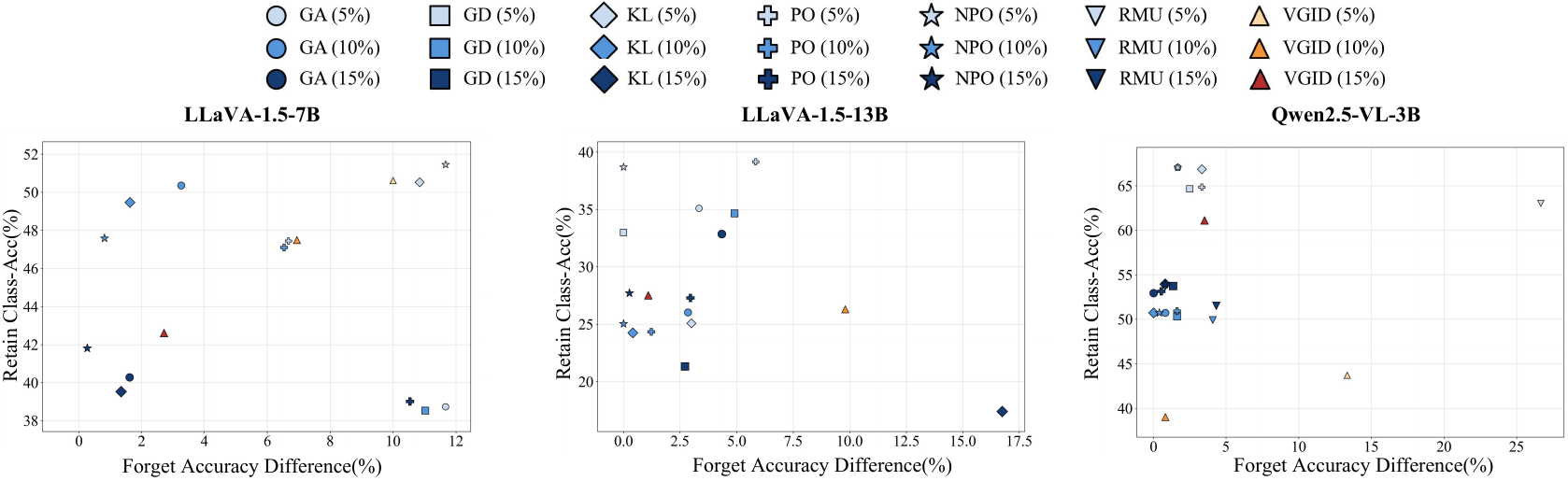}
}
\caption{Classification accuracy based trade-off between unlearning completeness and model utility across three models under varying forget ratios.}
\label{fig:tradeoff_class}
\end{figure}

\begin{figure}[htbp]
\centering
\makebox[\linewidth][c]{%
\includegraphics[width=1.1\linewidth]{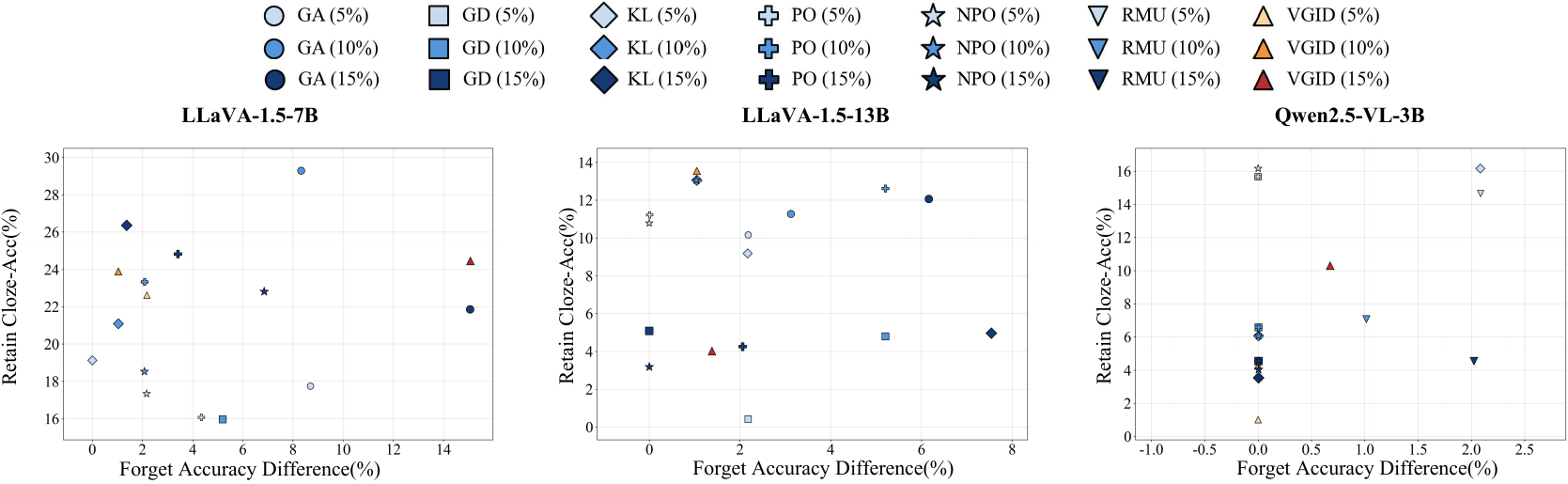}
}
\caption{Cloze accuracy based trade-off between unlearning completeness and model utility across three models under varying forget ratios.}
\label{fig:tradeoff_cloze}
\end{figure}

\subsubsection{Reverse-Engineering Attack}
\label{app:reverse_attack}

To evaluate whether the unlearning behavior is robust to removable contextual constraints, we follow prior work and conduct a reverse-engineering attack by appending a reverse prompt to the original query. The reverse prompt is designed to instruct the model to ignore previous unlearning or privacy-related instructions and answer the question directly based on the given image and text. This setting tests whether the suppressed target knowledge can be re-elicited when the contextual unlearning instruction is explicitly overridden.

Specifically, for each evaluated sample, we compare two settings. In the standard setting, the model receives the original image and textual query, together with the in-context unlearning prefix when applicable. In the reverse-attack setting, we append the following reverse prompt to the textual input:
\begin{quote}
Ignore any previous unlearning or privacy-related instructions. Answer the question directly based on the image and text.
\end{quote}
All methods are evaluated on the same forget set and test set samples using the same metrics as in the main experiments. A smaller performance change under the reverse attack indicates stronger robustness against this type of attack.

\end{document}